\documentclass[conference,table]{IEEEtran}
\IEEEoverridecommandlockouts
\usepackage{cite}
\usepackage{amsmath,amssymb,amsfonts}
\usepackage{algorithmic}
\usepackage{graphicx}
\usepackage{textcomp}
\usepackage{booktabs}
\usepackage{xcolor}
\def\BibTeX{{\rm B\kern-.05em{\sc i\kern-.025em b}\kern-.08em
    T\kern-.1667em\lower.7ex\hbox{E}\kern-.125emX}}

\usepackage[ruled,vlined]{algorithm2e}
\usepackage{amsthm}
\usepackage{mdframed}

\usepackage{multirow}
\usepackage{graphicx}

\usepackage{lipsum} % Chỉ để tạo văn bản giả, có thể bỏ đi nếu không cần
\usepackage{caption}
\usepackage{booktabs}
\usepackage{makecell}
\usepackage{standalone}

\usepackage[utf8]{inputenc}
\usepackage{subcaption}
\usepackage{colortbl}

\usepackage{pgfplots}
\pgfplotsset{compat=1.18} % Đặt phiên bản tương thích
\usepackage{tikz}
\usetikzlibrary{arrows.meta, positioning, shapes.geometric, arrows, fit, backgrounds, trees, mindmap}
\usepackage{multirow} % Gói hỗ trợ lệnh \multirow
\usepackage{forest}

\usepackage{array} % Preamble (nếu chưa thêm)

\newtheoremstyle{example}
  {3pt} {3pt}  % Không gian trên & dưới
  {}           % Font nội dung
  {}           % Indentation
  {\bfseries}  % Font tiêu đề
  {.}          % Dấu chấm sau tiêu đề
  { }          % Khoảng cách sau tiêu đề
  {}           % Header

\theoremstyle{example}
\newmdtheoremenv{example}{Example}

\begin{document}

\title{Speaking in Words, Thinking in Logic: A Dual-Process Framework in QA Systems\\
}

%\author{Anonymous Author(s)}

\author{\IEEEauthorblockN{1\textsuperscript{st} Tuan Bui}
\IEEEauthorblockA{\textit{Ho Chi Minh City University of Technology (HCMUT),}\\
\textit{Vietnam National University - Ho Chi Minh City} \\
Ho Chi Minh City, Vietnam\\
tuanbc88@hcmut.edu.vn}
\and
\IEEEauthorblockN{2\textsuperscript{nd} Trong Le}
\IEEEauthorblockA{\textit{Ho Chi Minh City University of Technology (HCMUT),}\\
\textit{Vietnam National University - Ho Chi Minh City} \\
Ho Chi Minh City, Vietnam\\
lvtrong.sdh231@hcmut.edu.vn}
\and
\IEEEauthorblockN{3\textsuperscript{rd} Phat Thai}
\IEEEauthorblockA{\textit{Ho Chi Minh City University of Technology (HCMUT),}\\
\textit{Vietnam National University - Ho Chi Minh City} \\
Ho Chi Minh City, Vietnam\\
phat.thaiquang2004@hcmut.edu.vn}
\and
\IEEEauthorblockN{4\textsuperscript{th} Sang Nguyen}
\IEEEauthorblockA{\textit{Ho Chi Minh City University of Technology (HCMUT),}\\
\textit{Vietnam National University - Ho Chi Minh City} \\
Ho Chi Minh City, Vietnam\\
sang.nguyen.imp21@hcmut.edu.vn}
\and
\IEEEauthorblockN{5\textsuperscript{th} Minh Hua}
\IEEEauthorblockA{\textit{Ho Chi Minh City University of Technology (HCMUT),}\\
\textit{Vietnam National University - Ho Chi Minh City} \\
Ho Chi Minh City, Vietnam\\
minh.hua2006@hcmut.edu.vn}
\and
\IEEEauthorblockN{6\textsuperscript{th} Ngan Pham}
\IEEEauthorblockA{\textit{Ho Chi Minh City University of Technology (HCMUT),}\\
\textit{Vietnam National University - Ho Chi Minh City} \\
Ho Chi Minh City, Vietnam\\
ngan.pham1409207@hcmut.edu.vn}
\and
\IEEEauthorblockN{7\textsuperscript{th} Thang Bui}
\IEEEauthorblockA{\textit{Ho Chi Minh City University of Technology (HCMUT),}\\
\textit{Vietnam National University - Ho Chi Minh City} \\
Ho Chi Minh City, Vietnam\\
bhthang@hcmut.edu.vn}
\and
\IEEEauthorblockN{8\textsuperscript{th} Tho Quan}
\IEEEauthorblockA{\textit{Ho Chi Minh City University of Technology (HCMUT),}\\
\textit{Vietnam National University - Ho Chi Minh City} \\
Ho Chi Minh City, Vietnam\\
qttho@hcmut.edu.vn}
}

\maketitle

\begin{abstract}
%This document is a model and instructions for \LaTeX. This and the IEEEtran.cls file define the components of your paper [title, text, heads, etc.]. *CRITICAL: Do Not Use Symbols, Special Characters, Footnotes,  or Math in Paper Title or Abstract.

Recent advances in large language models (LLMs) have significantly enhanced question-answering (QA) capabilities, particularly in open-domain contexts. However, in closed-domain scenarios such as education, healthcare, and law, users demand not only accurate answers but also transparent reasoning and explainable decision-making processes. While neural-symbolic (NeSy) frameworks have emerged as a promising solution—leveraging LLMs for natural language understanding and symbolic systems for formal reasoning—existing approaches often rely on large-scale models and exhibit inefficiencies in translating natural language into formal logic representations.

To address these limitations, we introduce \textbf{Text-JEPA} (Text-based Joint-Embedding Predictive Architecture), a lightweight yet effective framework for converting natural language into first-order logic (NL2FOL). Drawing inspiration from dual-system cognitive theory, Text-JEPA emulates System~1 by efficiently generating logic representations, while the Z3 solver operates as System~2, enabling robust logical inference. To rigorously evaluate the NL2FOL-to-reasoning pipeline, we propose a comprehensive evaluation framework comprising three custom metrics: \textit{conversion\_score}, \textit{reasoning\_score}, and \textit{Spearman\_rho\_score}, which collectively capture the quality of logical translation and its downstream impact on reasoning accuracy.

Empirical results on domain-specific datasets demonstrate that Text-JEPA achieves competitive performance with significantly lower computational overhead compared to larger LLM-based systems. Our findings highlight the potential of structured, interpretable reasoning frameworks for building efficient and explainable QA systems in specialized domains.
\end{abstract}


\section{Introduction}

Recent advancements in large language models (LLMs) have significantly reshaped the landscape of information retrieval and question answering (QA). For fact-based queries and open-domain tasks, retrieval-augmented generation (RAG) approaches have demonstrated strong performance by combining document retrieval with LLM-based generation. However, in closed-domain QA systems—such as those in education, law, or healthcare—where information is complex, domain-specific, and often drawn from heterogeneous sources, users demand not only accurate answers but also explicit reasoning, traceability, and interpretability. In such contexts, traditional LLMs often struggle to provide reliable and explainable outputs.

This challenge aligns with the dual-process theory of cognition, as articulated by \cite{Kahneman2011ThinkingFA}, which distinguishes between System 1 (fast, heuristic, intuitive) and System 2 (slow, analytical, logical). Effective QA systems, particularly in high-stakes domains, require reasoning capabilities akin to System 2. Echoing this, \cite{lecun2022path} emphasizes the necessity of integrating structured reasoning mechanisms to enable AI systems to generalize, infer, and verify beyond surface-level pattern matching.

To meet this demand, the field has increasingly turned to Neural-Symbolic (NeSy) approaches, which integrate deep learning and symbolic reasoning. These methods often involve converting natural language into formal logic representations—such as Propositional Logic (PL), First-Order Logic (FOL), or Temporal Logic (TL)—followed by inference using theorem provers or constraint solvers \cite{pan-etal-2023-logic, chen-etal-2023-nl2tl}. Systems like LINC \cite{olausson-etal-2023-linc} leverage LLMs as semantic parsers to generate formal logic, which is then verified by external symbolic engines such as Prover9 or ASP solvers. While these pipelines improve robustness, they often rely on iterative error correction and operate in a disjointed fashion—wherein the language model and symbolic components are loosely coupled. This separation not only impairs contextual understanding during logic correction but also limits the symbolic engine’s ability to influence upstream generation. Moreover, reliance on large-scale LLMs introduces scalability and cost concerns.

In this work, we advocate for the use of FOL as the target formalism due to its expressiveness, readability, and compatibility with downstream reasoning tools and semantic frameworks (e.g., Web Ontology Language—OWL, Resource Description Framework—RDF). To bridge the gap between natural language and symbolic logic in an efficient and coherent manner, we propose the Text-based Joint Embedding Predictive Architecture (Text-JEPA)—an end-to-end model for natural language to FOL (NL2FOL) conversion. For reasoning, we integrate the Z3 solver, a high-performance satisfiability modulo theories (SMT) engine, capable of validating logical inference under complex constraints. Its applicability to domains like law and education makes it a suitable backend for our architecture.

Our contributions are summarized as follows:
\begin{itemize}
    \item We introduce \textbf{Text-JEPA}, a novel architecture for NL2FOL conversion that jointly captures linguistic semantics and the structural constraints of formal logic representations.
    
    \item We propose a unified QA framework that incorporates the \textbf{Z3} solver, which not only performs logical reasoning for answer derivation but also serves as an automated evaluation tool for the generated logic.

    \item We are the first to systematically investigate and \textbf{quantify the correlation} between the quality of NL2FOL translation and downstream reasoning performance through three custom-designed metrics: \textit{conversion\_score}, \textit{reasoning\_score}, and \textit{Spearman\_rho\_score}.
    
    \item We conduct extensive empirical evaluations on a domain-specific dataset curated by experts, ensuring alignment with both linguistic and logical requirements. In addition to quantitative results, we provide detailed analyses on system interpretability and failure cases.
\end{itemize}

\section{Related Work}

\textbf{Neural-Symbolic Question Answering.} Recent work in NeSy systems has explored combining the linguistic flexibility of LLMs with the formal rigor of symbolic reasoning engines. This is particularly relevant for closed-domain QA, where answers must be both accurate and explainable. Approaches like LINC~\cite{olausson-etal-2023-linc}, Logic-LM~\cite{pan-etal-2023-logic}, and SATLM~\cite{10.5555/3666122.3668096} follow a \emph{pipeline architecture}, where LLMs first translate NL into formal logic (typically FOL or SAT), and symbolic tools like Prover9 or Z3 are subsequently used for reasoning. While these systems show promise, they often treat translation and reasoning as disjoint stages, leading to brittle performance when translation errors propagate into downstream reasoning. Moreover, their reliance on large-scale LLMs raises practical concerns regarding efficiency and scalability in real-world deployments.

\textbf{Structured Reasoning in LLM-based Systems.}
To address the limitations of pure pipeline models, recent studies have introduced structured reasoning frameworks. For instance, Faithful-CoT~\cite{lyu-etal-2023-faithful} proposes a two-stage reasoning chain that decomposes complex queries into symbolic and natural subcomponents. Similarly, Logic-LM and LLM-DA~\cite{wang2024large} aim to interleave symbolic solvers with LLM-generated logic. However, these models are typically limited to specific symbolic backends (e.g., Prolog, Datalog), and few support SMT-level reasoning as required in high-stakes domains like law or education. Furthermore, most of these methods offer limited introspection into how translation quality affects reasoning outcomes, leaving a critical evaluation gap in the NL2FOL pipeline.

\textbf{Evaluation and Interpretability.}
Despite growing interest in NL2FOL systems, few works have systematically analyzed the relationship between formal logic generation quality and downstream reasoning effectiveness. Most prior evaluations focus on accuracy or F1-score over reasoning outcomes, without isolating how conversion errors impact final answers. Additionally, logic explainability is often handled heuristically, without integrating formal proof extraction or automated solver validation. 

Our work addresses these limitations by introducing an end-to-end system—Text-JEPA—that couples formal conversion with SMT-based reasoning (Z3 slover) and introduces a principled evaluation framework grounded in three metrics: \texttt{Conv\_Score}, \texttt{Reason\_Score}, and \texttt{SRho\_Score}.

\section{Methodology}

This section introduces the overall methodology of our proposed QA framework, which integrates natural language understanding with formal logic reasoning to produce accurate and explainable responses. The central idea is to convert NL questions, facts, and rules into FOL representations, followed by logical inference via symbolic theorem proving. Our system is designed to support various types of reasoning tasks, such as verifying the truth value of a statement, retrieving entities that satisfy a condition, or evaluating whether a specific instance meets a given requirement. These capabilities enable the framework to handle a broad spectrum of queries in educational and knowledge-based domains.

To illustrate the system’s capabilities, we present a motivating example in the context of course enrollment reasoning (see Example~\ref{ex:course-enrollment}). This scenario demonstrates how our framework processes FOL-encoded rules and facts to answer diverse query types in a logically consistent and interpretable manner.

\begin{example}[Course Enrollment Reasoning]
\label{ex:course-enrollment}
\raggedright
\small
${}$\\
\textbf{Facts:}
\begin{align*}
& enrolled(Alice, cs101), \quad completed(Bob, cs101) \\
& completed(Charlie, cs101), \quad completed(Charlie, cs102)
\end{align*}
\textbf{Rules:}
\begin{align*}
\forall x \big( & enrolled(x, cs102) \Rightarrow completed(x, cs101) \big) \\
\forall x \big( & completed(x, cs102) \Rightarrow eligible\_ta(x) \big)
\end{align*}
%\vspace{0.5em}
\textbf{Queries:}
\begin{itemize}
    \item \textbf{(Q1)} \textbf{Truth-Value:} Is Alice’s enrollment in CS102 valid? $\rightarrow$ \textit{Answer:} \textit{False} \quad (She hasn't completed CS101)
    \item \textbf{(Q2)} \textbf{Entity Listing:} Who is eligible to be a teaching assistant? $\rightarrow$
    \textit{Answer:} \textit{Charlie}
    \item \textbf{(Q3)} \textbf{Conditional Evaluation:} Can Bob enroll in CS102? $\rightarrow$
    \textit{Answer:} \textit{True} \quad (He completed CS101)
\end{itemize}
\end{example}

\begin{figure}[h]
    \centering
    \includegraphics[width=0.5\textwidth]{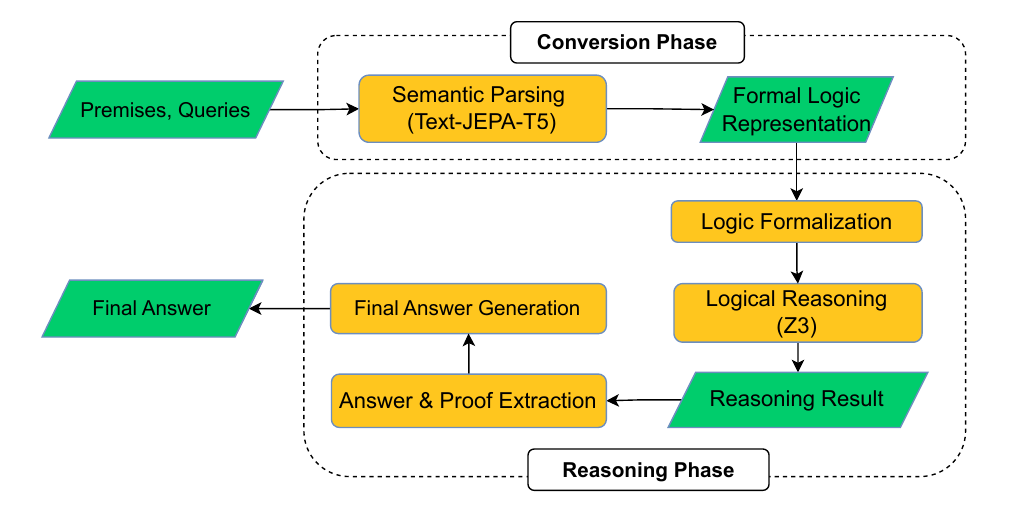}
    \caption{System architecture: Natural language inputs are converted to formal logic by Text-JEPA, followed by logical inference using Z3.}
    \label{fig_architecture}
\end{figure}

The overall architecture of our QA system consists of two main stages, as shown in Figure~\ref{fig_architecture}:

\begin{enumerate}
    \item \textbf{Conversion Phase:} Natural language inputs (rules, facts, and queries) are converted into FOL expressions using the Text-based Joint Embedding Predictive Architecture (Text-JEPA).
    
    \item \textbf{Reasoning Phase:} The FOL representations are translated into Z3-compatible syntax and processed using a theorem prover to derive logical outcomes.
\end{enumerate}

\subsection{Conversion Phase}

\begin{figure*}[!t]
    \centering
    \includegraphics[width=0.9\textwidth]{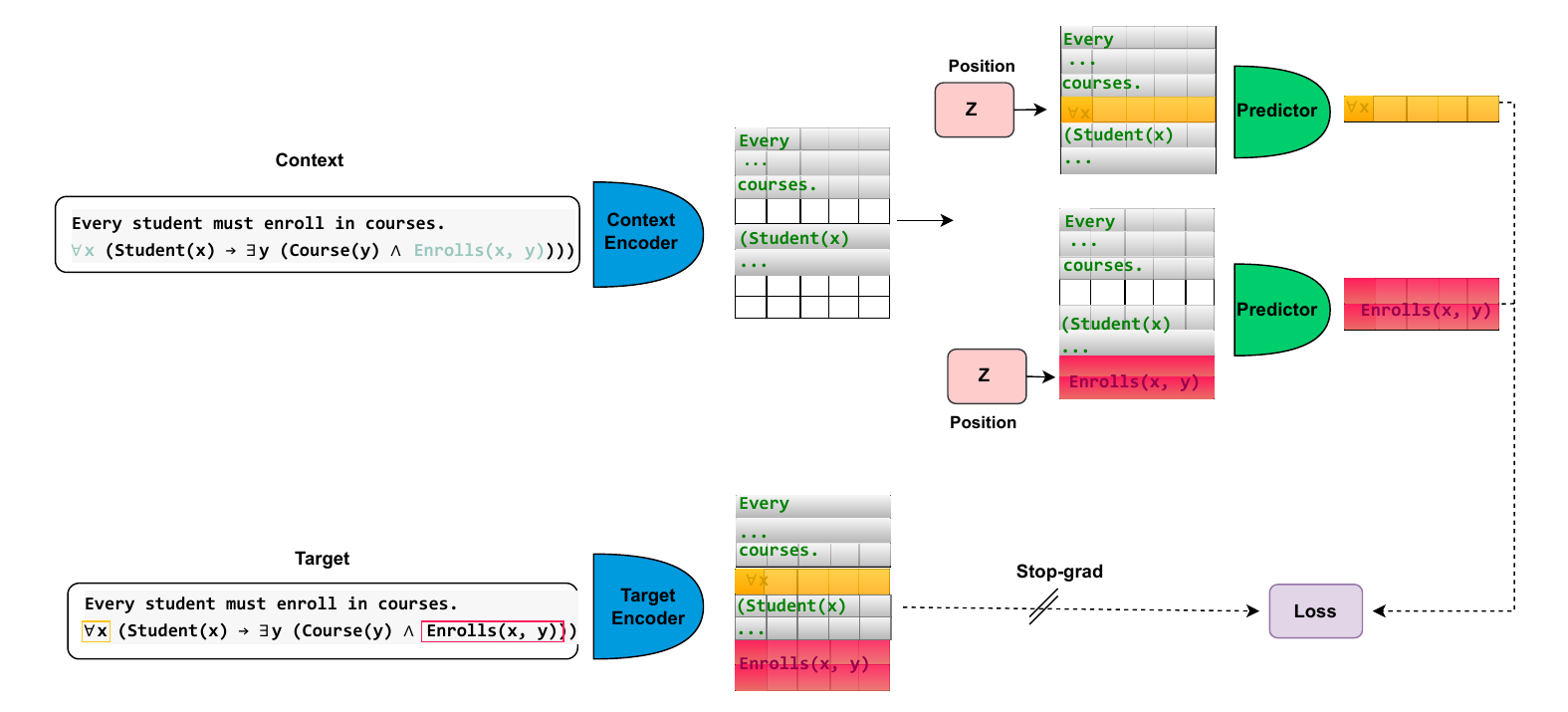}
    \caption{An overview of the \textbf{Text-JEPA} architecture for learning mappings between NL and FOL. The system encodes input sentences and their FOL forms using a context encoder, identifies key positions (e.g., for $\forall x$ and $Enrolls(x, y)$), and predicts masked components via dual predictors. A target encoder and a loss function guide training by comparing predicted and target representations, enabling the model to capture fine-grained semantic alignments.}
    \label{fig_architecture_text_jepa}
\end{figure*}

We propose \textbf{Text-JEPA} (Text-based Joint Embedding Predictive Architecture), an encoder-decoder architecture inspired by JEPA~\cite{lecun2022path}, specifically fine-tuned for NL2FOL conversion. Figure~\ref{fig_architecture_text_jepa} provides an overview of the \textbf{Text-JEPA} architecture for learning the mapping between natural language (NL) and First-Order Logic (FOL). Text-JEPA comprises the following components:

% \begin{itemize}
%     \item \textbf{Target Encoder:} Encodes the input text into dense vector representations capturing semantic content.
%     \item \textbf{Context Encoder:} Integrates contextual cues (e.g., task type, surrounding entities) to aid logical form prediction.
%     \item \textbf{Predictor Module:} Generates or refines logical forms by masked prediction and partial structure recovery.
% \end{itemize}

\begin{itemize}
    \item \textbf{Targets:} Combines natural language and logical sentences into a single sequence, tokenizes and groups them into clusters to create representations, and randomly samples blocks to encode contextual and structural relationships.
    \item \textbf{Context:} Removes target block tokens from the sequence, encodes the remaining tokens to capture semantic structure, and provides a foundation for predicting missing representations.
    \item \textbf{Predictor:} Reconstructs target block representations using the context output, employs positional mask tokens to maintain spatial dependencies, and iteratively predicts all target blocks for coherent reconstruction.
\end{itemize}

% To improve robustness, we further fine-tuned the model with noisy data. Training involves synthetic perturbations. Training involves synthetic perturbations:

% \begin{itemize}
%     \item Altering logical operators or quantifiers,
%     \item Swapping variable/predicate names,
%     \item Introducing syntactic or semantic inconsistencies.
% \end{itemize}

% This denoising strategy enables the model to enhance the generation of accurate FOLs.

\subsection{Reasoning Phase}

The reasoning phase operates on well-formed FOL representations derived from natural language input. Given a set of premises (facts and rules) and a query (hypothesis or question), the system first parses the FOL expressions into structured syntax trees and normalizes their components, including constants, variables, predicates, logical connectives, and quantifiers. To ensure consistency across the knowledge base, a \textit{predicate alignment} step unifies semantically equivalent predicates (e.g., \texttt{Student(x)} vs. \texttt{Students(x)}).

The normalized formulas are then encoded into Z3 syntax and added to a solver context. The query is negated and also submitted to the solver to test for logical entailment via satisfiability checking. If the result is \texttt{unsat}, the query is logically entailed by the premises; if \texttt{sat}, a counterexample model is extracted, indicating that the entailment does not hold.

To support interpretability, the system extracts either a proof or a counterexample depending on the inference outcome. This information is passed to an Answer Generation Module, which produces a natural language explanation summarizing the logical reasoning process and final conclusion.

Algorithm~\ref{algorithm_1} outlines the complete pipeline, from symbolic parsing and alignment to Z3-based reasoning and natural language explanation.

\begin{algorithm}[h]
\caption{Logic-Based QA via FOL Parsing and Z3 Reasoning}
\label{algorithm_1}
\small
\KwIn{An input instance consisting of: A set of \texttt{FOL\_text} premises (facts and rules), and queries.}
\KwOut{Answer indicating whether the query is logically entailed by the premises, along with optional explanations.}

\vspace{0.5em}

\textbf{Step 1 -- Data Loading:} Load the \texttt{FOL\_text} fields, which include formal FOL representations of premises and queries. \\

\textbf{Step 2 -- Structure Parsing \& Predicate Alignment:} \\
Decompose FOL into formal components including constants, variables, predicates, logical connectives, and quantifiers. Normalize into structured syntax trees. \\
\textit{Predicate Alignment:} Unify semantically equivalent predicates (e.g., `Student(x)` vs. `Students(x)`) to ensure consistent symbolic representation across premises and queries. \\

\textbf{Step 3 -- Z3 Encoding:} Translate all FOL premises into Z3 expressions and add them to a solver context via \texttt{s.add(...)}. Negate the query or hypothesis and also add it to the solver. \\

\textbf{Step 4 -- Logical Reasoning:} Invoke \texttt{s.check()} to perform logical inference: \\
\hspace{1em} $\bullet$ If the result is \texttt{unsat}, the query logically follows from the premises.\\
\hspace{1em} $\bullet$ If \texttt{sat}, the query is not entailed; a counterexample (model) is extracted. \\

\textbf{Step 5 -- Answer \& Proof Extraction:} Based on the result from Step 4, extract either the proof (for \texttt{unsat}) or a counterexample model (for \texttt{sat}). This forms the basis for explanation and justification. \\

\textbf{Step 6 -- Final Answer Generation:} Use the extracted information (proof or counterexample) to generate a human-readable answer and explanation, indicating why the query is valid or invalid.

\normalsize
\end{algorithm}

The example above demonstrates the application of our reasoning framework to verify the logical validity of candidate conclusions against a given rule. The natural language question explores whether a student without any MOS certificates can still be allowed to undertake a graduation thesis. The formalized premise explicitly states that a student must possess all three certificates—Word, Excel, and PowerPoint—in order to qualify for the thesis.

Three candidate conclusions are proposed to challenge the entailment: a generic assertion without conditions (1), a case of complete absence of certificates (2), and a scenario where only one certificate is missing (3). Each of these hypotheses is translated into First-Order Logic and submitted to the Z3 solver for verification.

\begin{example}[Algorithm~\ref{algorithm_1}]
\raggedright
\small
${}$\\
\textbf{NL Question:} \\
Is someone allowed to do a graduation thesis even if they haven’t earned any MOS certificates? \\[0.8em]

\textbf{Premise:} \\
\hspace{0.2em} A student is allowed to do a graduation thesis only if they have all three MOS certificates: Word, Excel, and PowerPoint. \\ [0.3em]
\hspace{0.2em} $\forall x ( student(x) \land has\_cert(x, \text{word}) \land has\_cert(x, \text{excel}) \land has\_cert(x, \text{powerpoint}) \Rightarrow do\_thesis(x) )$ \\[0.8em]

\textbf{Candidate Conclusions:} \\
\hspace{0.2em} (1) $do\_thesis(x)$ \\
\hspace{0.2em} (2) $\neg has\_cert(x, \text{word}) \land \neg has\_cert(x, \text{excel}) \land \neg has\_cert(x, \text{powerpoint}) \Rightarrow do\_thesis(x)$ \\
\hspace{0.2em} (3) $student(x) \land \neg has\_cert(x, \text{excel}) \Rightarrow do\_thesis(x)$ \\[0.8em]

\textbf{Parsed Elements:} \\
\hspace{0.2em} \textit{Variables:} $x$ \\
\hspace{0.2em} \textit{Constants:} ``word'', ``excel'', ``powerpoint'' \\
\hspace{0.2em} \textit{Predicates:} $student(x)$, $has\_cert(x, \_)$, $do\_thesis(x)$ \\
\hspace{0.2em} \textit{Connectives:} $\land$, $\Rightarrow$, $\neg$ \quad \\
\hspace{0.2em} \textit{Quantifier:} $\forall x$ \\[0.8em]

\textbf{Matched Logical Form:} \\
\hspace{0.2em} $student(x) \land \neg has\_cert(x, \text{word}) \land \neg has\_cert(x, \text{excel}) \land \neg has\_cert(x, \text{powerpoint}) \Rightarrow do\_thesis(x)$ \\[0.8em]

\textbf{Z3 Output:} \\
\hspace{0.2em} \texttt{sat} — A counterexample exists: a student without any certificates who is not allowed to do the thesis. \\[0.3em]

\textbf{Final Answer: False}
\end{example}

In this case, Z3 returns \texttt{sat}, indicating that the candidate logical form (3) is not entailed by the premise. A counterexample is found, confirming that the conclusion does not logically follow. This showcases the system’s ability to detect invalid reasoning patterns, even in nuanced cases where some—but not all—prerequisites are satisfied.

In summary, our framework offers a modular and explainable pipeline from natural language inputs to logical conclusions. It balances the flexibility of neural models with the rigor of symbolic inference, making it well-suited for domain-specific QA applications where transparency and correctness are critical.

%---------------------------------------------
\section{Evaluation Metrics}

To comprehensively assess the performance of our NL2FOL and reasoning pipeline, we employ three principal evaluation metrics: \textbf{Conversion Score} (\texttt{conv\_score}), \textbf{Reasoning Score} (\texttt{reason\_score}), and \textbf{Spearman Rank Correlation} (\texttt{srho\_score}). These metrics are designed to evaluate both the structural and semantic fidelity of FOL translations, as well as the effectiveness of downstream logical reasoning.

\subsection{Conversion Score}

We define a composite Conversion Score to quantify the quality of the NL2FOL conversion. It integrates three core dimensions:

\begin{itemize}
    \item \textbf{Syntactic Well-formedness (SWF)}: Measures whether the generated FOL adheres to the formal syntax of logic. Each FOL expression is evaluated against six syntactic criteria (Table~\ref{tab:fol_standards}), with a binary score assigned to each. The final score is the average across all criteria.
    
    \item \textbf{Predicate-level Semantic Equivalence (PSE)}: Assesses the degree of alignment between predicates and entities in the predicted and ground-truth FOL. Cosine similarity is computed between predicate embeddings after bidirectional alignment, and the average similarity over matched and unmatched predicates is taken as the PSE score.

    \item \textbf{Logical Equivalence (LE)}: Evaluates whether the predicted and gold FOL formulas are logically equivalent by comparing their truth tables. The LE score is defined as the proportion of matching rows across both truth tables, following the approach of \cite{yang-etal-2024-harnessing}.
\end{itemize}

\begin{table}[h]
    \centering
    \caption{Criteria for evaluating Syntactic well-formedness}
    \renewcommand{\arraystretch}{1.3}
    \begin{tabular}{p{1.2cm} p{6.4cm}}
        \toprule
        \textbf{Standards} & \textbf{Description} \\
        \midrule
        \makecell{\textbf{Variable}} & Variables in predicates must only contain lowercase alphabetical characters. \\
        & \textbf{Example:} $\forall x\ (\text{Blake}(x \to \text{Building}(x)))$ \newline $\to$ Misplaced parenthesis. \\
        \hline
        \makecell{\textbf{Syntactic}} & Every variable must be defined before being used. \\
        & \textbf{Example:} $\text{Luxury}(x) \to \text{Shopping}(x)$ \newline $\to$ $x$ is undefined. \\
        \hline
        \multirow{2}{*}{ \makecell{\textbf{Semantic} \\ \textbf{Validity}} }
         & Logical operators must be formed correctly to express a valid operation. \\
        & \textbf{Example:} $\text{Wake}(\text{hulk}) \to\to \text{BreakBridge}(\text{hulk})$ \newline $\to$ Invalid operator usage. \\
        \hline
        \multirow{2}{*}{ \makecell{\textbf{Parentheses} \\ \textbf{Validity}} }& Parentheses must be opened and closed correctly. \\
        & \textbf{Example:} $\text{Code}(x) \wedge \text{Mac}(x))$  \newline $\to$ Extra closing parenthesis. \\
        \hline
        \multirow{2}{*}{\makecell{\textbf{Comparison} \\ \textbf{Symbols}}} & FOL must not contain comparison symbols such as $>, <, =$. \\
        & \textbf{Example:} $\text{Height}(x) > \text{Weight}(x)$ \newline $\to$ Contains $>$. \\
        \hline
        \multirow{2}{*}{  \makecell{\textbf{Special} \\ \textbf{characters}}} & FOL can only contain English alphabetical characters and logical operators. \\
        & \textbf{Example:} $\forall x\ (\text{Reads}(x) \to \text{Gain}?(x))$ \newline $\to$ Contains invalid $?$. \\
        \bottomrule
    \end{tabular}
    \label{tab:fol_standards}
\end{table}

We combine these components into a single conversion score as follows:
\begin{equation}
\texttt{conv\_score} = \lambda_1 \cdot \frac{2 \cdot \text{SWF} \cdot \text{LE}}{\text{SWF} + \text{LE}} + \lambda_2 \cdot \text{PSE}
\end{equation}
where $\lambda_1 + \lambda_2 = 1$. This formulation emphasizes the necessity of both syntactic correctness and logical validity for reasoning, while also considering semantic fidelity.

\subsection{Reasoning Score}

The Reasoning Score quantifies the accuracy of automated inference using the converted FOL. Given the gold entailment label and the predicted outcome from a reasoner, each instance is scored based on the following rule:

\begin{equation}
\texttt{reason\_score}_i =
\begin{cases} 
    S_{\max}, & \text{if } R_i = G_i \\ 
    S_{\text{mid}}, & \text{if } R_i \neq G_i \text{ and executable} \\  
    S_{\min}, & \text{if } R_i = \text{error\_compile}
\end{cases}
\end{equation}

where $R_i$ and $G_i$ denote the predicted and gold reasoning results, respectively. We set $S_{\max}=1$, $S_{\text{mid}}=0.5$, and $S_{\min}=0$, corresponding to full correctness, partial credit for logically valid but incorrect predictions, and zero credit for syntactic failures.

\subsection{SRho\_Score: Spearman Rank Correlation}

To evaluate the relationship between FOL conversion quality and downstream reasoning performance, we compute the Spearman rank correlation coefficient, denoted as \texttt{srho\_score}, between \texttt{conv\_score} and \texttt{reason\_score}. It is calculated as follows:

\begin{equation}
\texttt{srho\_score} = 1 - \frac{6 \sum d_i^2}{n(n^2 - 1)}
\end{equation}

where $d_i$ is the difference in ranks between the conversion and reasoning scores for instance $i$, and $n$ is the total number of evaluated samples.

This metric offers insight into how closely the structural and semantic quality of generated FOL formulas impacts the effectiveness of logical reasoning. Interpretation of the score is as follows:
\begin{itemize}
    \item \texttt{srho\_score} $\approx 1$: Strong positive correlation, indicating that better FOL conversion tends to yield more accurate reasoning outcomes.
    \item \texttt{srho\_score} $\approx 0$: Weak or no correlation, suggesting that FOL conversion quality has little to no effect on reasoning accuracy.
    \item \texttt{srho\_score} $< 0$: Negative correlation, potentially indicating incompatibility or misalignment between FOL structure and the reasoning engine.
\end{itemize}

This correlation-based analysis serves as a diagnostic tool for evaluating both the fidelity of FOL generation and its practical utility in automated reasoning pipelines.

\section{Experimentation}

\subsection{Experimental setup}

\paragraph{Datasets} We evaluate our method on two benchmark datasets for NL2FOL: \textbf{MALL} \cite{yang-etal-2024-harnessing} and \textbf{FOLIO}\cite{han-etal-2024-folio}. The MALL dataset is used for supervised training and in-domain evaluation, following an 80\%-20\% train-test split. To assess generalization, we additionally evaluate on FOLIO, which is fully out-of-domain and contains natural language statements paired with gold FOL representations and entailment labels for reasoning.

\paragraph{Models} We compare our proposed \textbf{Text-JEPA-T5-base} model against \textbf{T5-base}, \textbf{Gemini 1.5 Flash} and \textbf{Gemini 1.0 Pro}. All models are evaluated under zero-shot (ZS) and 10-shot (10S) settings. % For our model, few-shot examples are selected based on semantic similarity to the input.

\paragraph{Conversion Evaluation.} We evaluate the NL2FOL conversion stage using four metrics: syntactic well-formedness (SWF), predicate-level semantic equivalence (PSE), logical equivalence (LE), and a composite conversion score (\textbf{Conv-Score}) that aggregates the three.

\paragraph{Reasoning Evaluation}
We evaluate reasoning as a 3-way entailment classification task, where the ground-truth conclusion\_label takes one of {true, false, uncertain} (\textbf{Accuracy-Score}). To handle conversion issues, we allow the system's prediction, conclusion\_predict, to optionally include a fourth value: compile\_error, indicating that the generated FOL is syntactically invalid or not executable. This setup enables joint evaluation of both reasoning ability and conversion robustness. We compute reasoning accuracy (\textbf{Reason-Score}) on the FOLIO dataset using both symbolic (Z3) and neural (Gemini) reasoners. We also report the \textbf{SRho-Score}, the Spearman correlation between Conv-Score and Reason-Score, to assess how conversion quality influences reasoning performance.

\subsection{Experimental results}

Table~\ref{tab:conversion_stage_combined} reports the performance of different models and prompting strategies on the NL-to-FOL conversion task across two datasets: MALL (in-domain) and FOLIO (out-of-domain). On the in-domain MALL dataset, all models achieve relatively high performance. The fine-tuned T5-base model performs competitively, achieving a Conv-Score of 85.30 with 10-shot prompting. However, Text-JEPA outperforms both Gemini and fine-tuned T5-base across all individual metrics (SWF, PSE, and LE), suggesting better alignment between syntactic form and underlying logical meaning. While T5-base does well in SWF, its lower PSE and LE indicate weaker predicate grounding and logical consistency, underscoring the benefit of Text-JEPA’s joint semantic-logical modeling.

On the out-of-domain FOLIO dataset, the advantages of Text-JEPA become more pronounced. It significantly outperforms all Gemini variants, achieving a Conv-Score of 82.89 in the 10S (Rank1) setting compared to Gemini’s best score of 63.93. Moreover, Text-JEPA is robust to prompt variation, showing only a minor drop from 10S (Max) to 10S (Avg), whereas Gemini suffers a steep performance decline. Interestingly, Gemini's ZS and 10S (Max) variants yield identical results, implying the API may employ internal few-shot or retrieval mechanisms. Overall, despite being built on the lightweight T5-base architecture, Text-JEPA demonstrates superior generalization and task alignment without relying on massive proprietary models.

\begin{table}[t]
    \centering
    \renewcommand{\arraystretch}{1.4} % tăng chiều cao dòng
    \caption{Comparison of methods in terms of SWF, PSE, and LE on MALL (in-domain) and FOLIO (out-of-domain). Results are reported for three models: \textbf{T5-base (Fine-tuned)}, \textbf{Gemini 1.5 Flash} and our proposed model, \textbf{Text-JEPA}.}

    \label{tab:conversion_stage_combined}
   % \resizebox{\columnwidth}{!}{%
    \begin{tabular}{|l|l|c|r|r|r|r|}
        \hline
        \textbf{DS.} & \textbf{Model} & \textbf{Method}  & \textbf{SWF} & \textbf{PSE} & \textbf{LE} & 
\makecell{\textbf{Conv} \\ \textbf{-Score}} \\
        \hline
        \multirow{4}{*}{ \rotatebox{90}{MALL}} 
        & Gemini & ZS      & 76.3 & 52.8 & 60.6 & 60.1 \\
        & Gemini & 10S & 78.2 & 60.5 & 67.1 & 66.3 \\
        \cline{2-7}
        & T5-base & ZS      & 87.40 & 72.20 & 83.50 & 78.80 \\
        & T5-base & 10S & 88.00 & 82.80 & 87.70 & 85.30 \\
        \cline{2-7}
        & \textbf{Text-JEPA} & ZS      & \textbf{88.7} & \textbf{77.2} & \textbf{86.6} & \textbf{82.4} \\
        & \textbf{Text-JEPA} &  10S & \textbf{89.1} & \textbf{85.4} & \textbf{88.8} & \textbf{87.1} \\
        \hline
        \multirow{4}{*}{\rotatebox{90}{FOLIO}} 
        & Gemini & ZS      & 92.98 & 54.51 & 71.28 & 63.93 \\
        & Gemini & 10S (Max)& 92.98 & 54.51 & 71.28 & 63.93 \\
        & Gemini & 10S (Avg)& 84.59 & 47.10 & 54.56 & 51.53 \\
       
        \cline{2-7}
        & \textbf{Text-JEPA} & ZS      & \textbf{97.58} & \textbf{61.49} & \textbf{83.85} & \textbf{74.01} \\
        & \textbf{Text-JEPA} & 10S (Max) & \textbf{98.29} & \textbf{85.84} & \textbf{63.52} & \textbf{75.44} \\
        & \textbf{Text-JEPA} & 10S (Avg) & \textbf{97.91} & \textbf{84.40} & \textbf{62.33} & \textbf{74.64} \\
        & \textbf{Text-JEPA} & 10S (Rank1) & \textbf{99.21} & \textbf{75.12} & \textbf{88.47} & \textbf{82.89} \\
        \hline
    \end{tabular}
    %}
\end{table}

\begin{table*}[t]
    \centering
    \small
    \renewcommand{\arraystretch}{1.5}
    \caption{Evaluation of different methods for NL2FOL conversion and reasoning on the FOLIO dataset.}
    \label{tab:conversion_reasong_on_FOLIO}
    \begin{tabular}{|
        >{\centering\arraybackslash}p{0.5cm}|  % No
        >{\centering\arraybackslash}p{1.2cm}|  % model
        >{\centering\arraybackslash}p{2.0cm}|  % method
        >{\centering\arraybackslash}p{1.3cm}|
        >{\centering\arraybackslash}p{1.9cm}|
         >{\centering\arraybackslash}p{1.3cm}|
        >{\centering\arraybackslash}p{1.3cm}|
        >{\centering\arraybackslash}p{1.3cm}|}
        \hline
        \multicolumn{4}{|c|}{\textbf{Conversion Stage}} & \multicolumn{4}{c|}{\textbf{Reasoning Stage}} \\
        \hline
        \textbf{Case} & \textbf{Model} & \textbf{Method} & \textbf{Conv-Score} & \textbf{Reasoner} & \textbf{Accuracy} & \textbf{Reason-Score} & \textbf{SRho-Score} \\
        \hline
        1.1 & \multirow{2}{*}{\makecell{Gemini \\ 1.0\\Pro}}  & \multicolumn{2}{c|}{Direct on NL} & Gemini (ZS) & 53.4 & --   &  -- \\
        \cline{1-1} \cline{3-8}
        1.2 &  & \multicolumn{2}{c|}{Direct on NL} & Gemini (CoT) & 57.35& --  & -- \\
        
        \hline
        2.1 & \multirow{3}{*}{\makecell{Gemini \\ 1.5\\Flash}}  & \multicolumn{2}{c|}{Direct on NL} & Gemini (ZS) & \textbf{65.1} & --  & -- \\
        \cline{1-1} \cline{3-8}
        2.2 &  & \multicolumn{2}{c|}{Direct on NL} & Gemini (CoT) & 63.2 & --  & -- \\
        \cline{1-1} \cline{3-8}
        2.3 & \multirow{2}{*}{--}  & \multicolumn{2}{c|}{Direct on FOL\_label} & Gemini (ZS) & 62.1 & --  & -- \\

        \hline
        3.1 & \multirow{4}{*}{\makecell{Gemini \\ 1.5\\Flash}} & \multirow{2}{*}{\makecell{ZS}} & \multirow{2}{*}{\makecell{63.93}}  & Gemini (ZS) & 42.16 & 71.08 & 0.052 \\
        \cline{1-1} \cline{5-8}
        3.2 &  &     &      & Z3 & 21.08 & 38.97 & \textbf{0.466} \\
        
        \cline{1-1} \cline{3-8}
        3.3 &  & \multirow{2}{*}{\makecell{CoT}}  & \multirow{2}{*}{\makecell{66.25}}  & Gemini (CoT) & 42.16 & 71.08 & 0.079 \\
        \cline{1-1} \cline{5-8}
        3.4 &  &     &      & Z3 & 22.55 & 41.18 & 0.406 \\
        
        \hline
        4.1 & \multirow{3}{*}{\makecell{\textbf{Text} \\ \textbf{-JEPA} \\ \textbf{T5-base} \\ \textbf{(Ours)} }} & ZS & 74.01 & Z3 & 30.39 & \textbf{58.55} & 0.192 \\
        \cline{1-1} \cline{3-8}
        4.2 & & 10S (Rank 1) & \textbf{82.89} & Z3 & 30.39 & 55.63 & 0.258 \\
        \cline{1-1} \cline{3-8}
        4.3 &  & \multicolumn{2}{c|}{Direct on FOL\_label} & Z3 & \textbf{95.0} & --  & -- \\
        \hline
    \end{tabular}
\end{table*}

Table~\ref{tab:conversion_reasong_on_FOLIO} presents an in-depth evaluation of end-to-end reasoning pipelines on the FOLIO dataset. The results are analyzed across four distinct settings based on model and reasoning strategies.

Case 1 investigates Gemini 1.0 Pro performing direct reasoning on natural language (NL) inputs. Results show moderate performance, with accuracy scores of 53.4 (zero-shot) and 57.35 (chain-of-thought), indicating limited reasoning capabilities when applied directly to unstructured inputs.

Case 2 involves Gemini 1.5 Flash, which demonstrates clear improvements over Gemini 1.0 Pro under similar conditions. Accuracy increases to 65.1 (ZS) and 63.2 (CoT), suggesting that Gemini has undergone meaningful upgrades across versions. Interestingly, when Gemini 1.5 Flash reasons over FOL\_label (structured logical forms), the performance remains comparable (62.1), only slightly below its NL reasoning counterpart. This implies that Gemini's newer version is capable of extracting similar reasoning performance regardless of input format, possibly due to stronger latent alignment mechanisms.

Case 3 explores a two-stage pipeline using Gemini 1.5 Flash for both conversion and reasoning. In Case 3.1, Gemini (ZS) is used for both stages, achieving high accuracy (42.16) and Reason-Score (71.08), but a very low SRho-Score (0.052), revealing that the model likely relies on implicit, uninterpretable patterns in reasoning rather than aligning with the underlying logical structure. In contrast, Case 3.2 uses the Z3 symbolic reasoner over Gemini-generated formulas. While this yields lower accuracy (21.08) and Reason-Score (38.97), it results in a substantially higher SRho (0.466), indicating much stronger logical faithfulness between intermediate formulas and final predictions. This trade-off underscores the opacity of LLM-based reasoners and the value of symbolic tools for traceable inference.

Case 4 evaluates our proposed model, Text-JEPA, in conjunction with symbolic reasoning via Z3. Despite Text-JEPA achieving superior conversion quality (Conv-Score up to 82.89) and competitive Reason-Scores (e.g., 55.63), the corresponding SRho-Scores are modest (max 0.258). We attribute this to a design difference: our model performs conversion at the sentence level, whereas Gemini operates at paragraph level, resulting in reduced continuity and inter-sentence context when reasoning. Lastly, Case 4.3 demonstrates an upper bound for structured input reasoning, with Z3 applied directly to gold FOL\_labels, achieving an impressive 95.0\% accuracy, validating the utility of symbolic reasoners when given perfect logical forms.

\begin{figure}[h!]
    \centering
    \scalebox{0.54}{%
        \begin{minipage}[t]{0.95\textwidth}
        
            % First Row
            \begin{subfigure}[t]{0.5\textwidth}
                \centering
                \includegraphics[width=\linewidth]{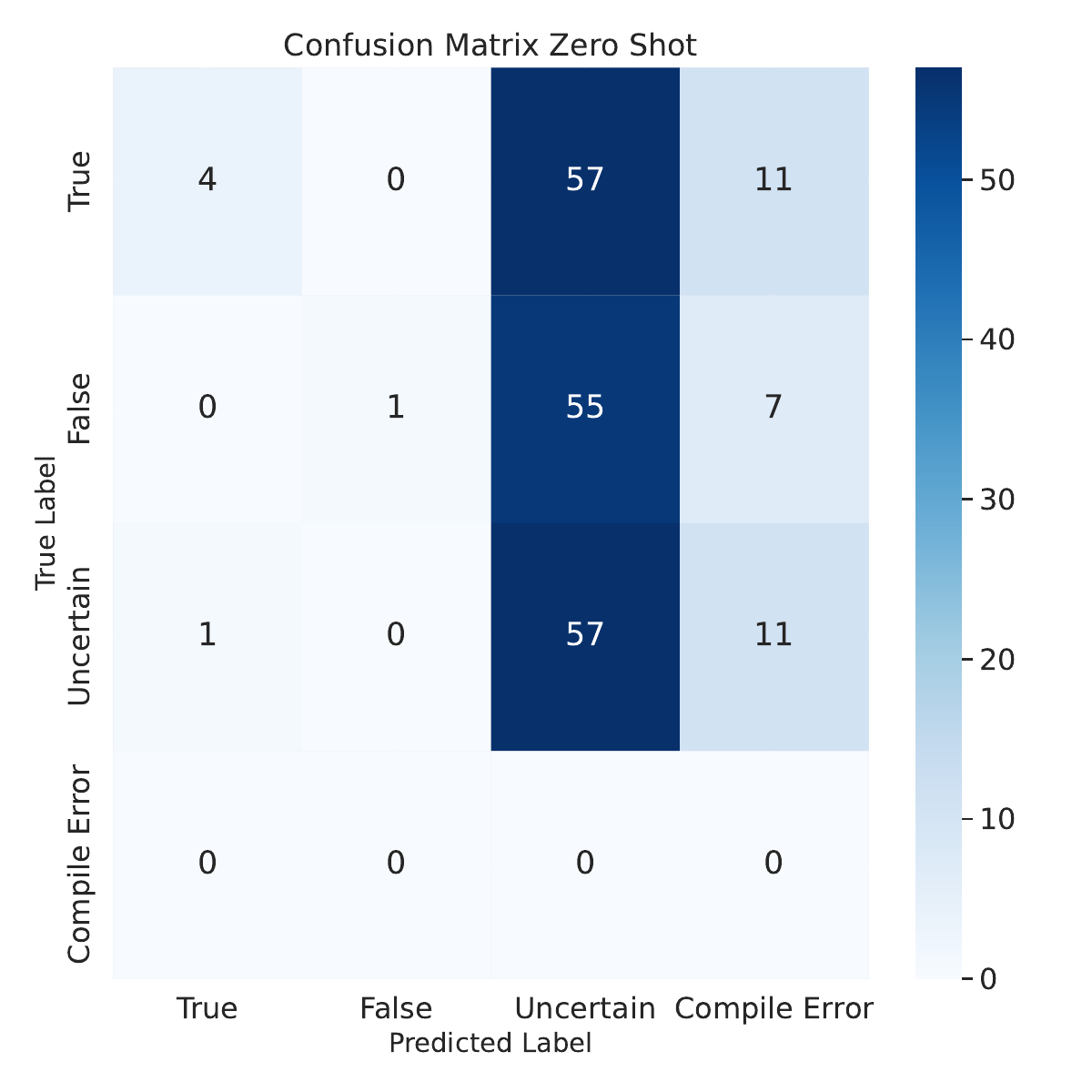}
                \caption{Zero Shot}
            \end{subfigure}
            \hfill
            \begin{subfigure}[t]{0.5\textwidth}
                \centering
                \includegraphics[width=\linewidth]{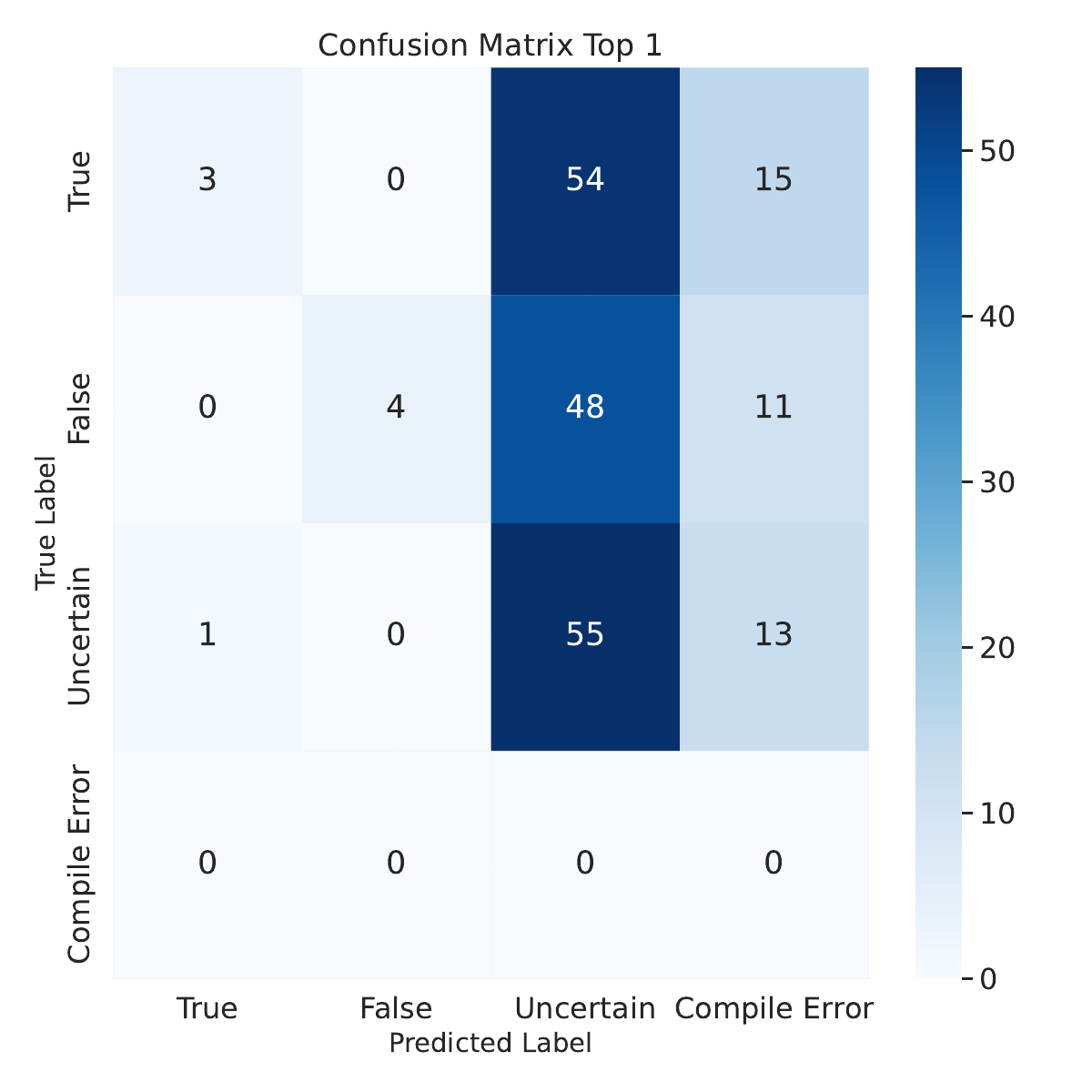}
                \caption{10-Shot \& Rank 1}
            \end{subfigure}
            
            \vspace{1em}
            
            % Second Row
            \begin{subfigure}[t]{0.5\textwidth}
                \centering
                \includegraphics[width=\linewidth]{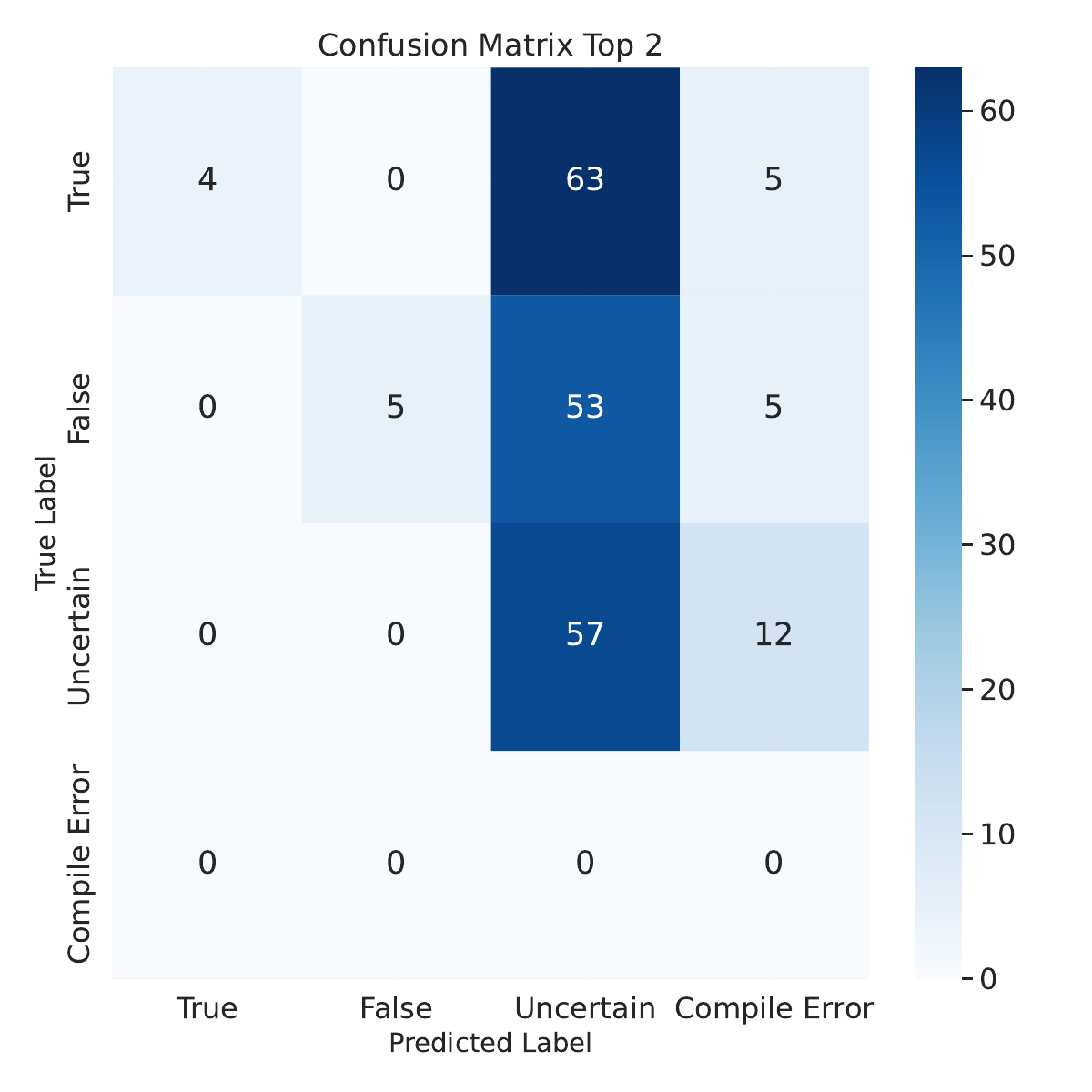}
                \caption{10-Shot \& Top 2}
            \end{subfigure}
            \hfill
            \begin{subfigure}[t]{0.5\textwidth}
                \centering
                \includegraphics[width=\linewidth]{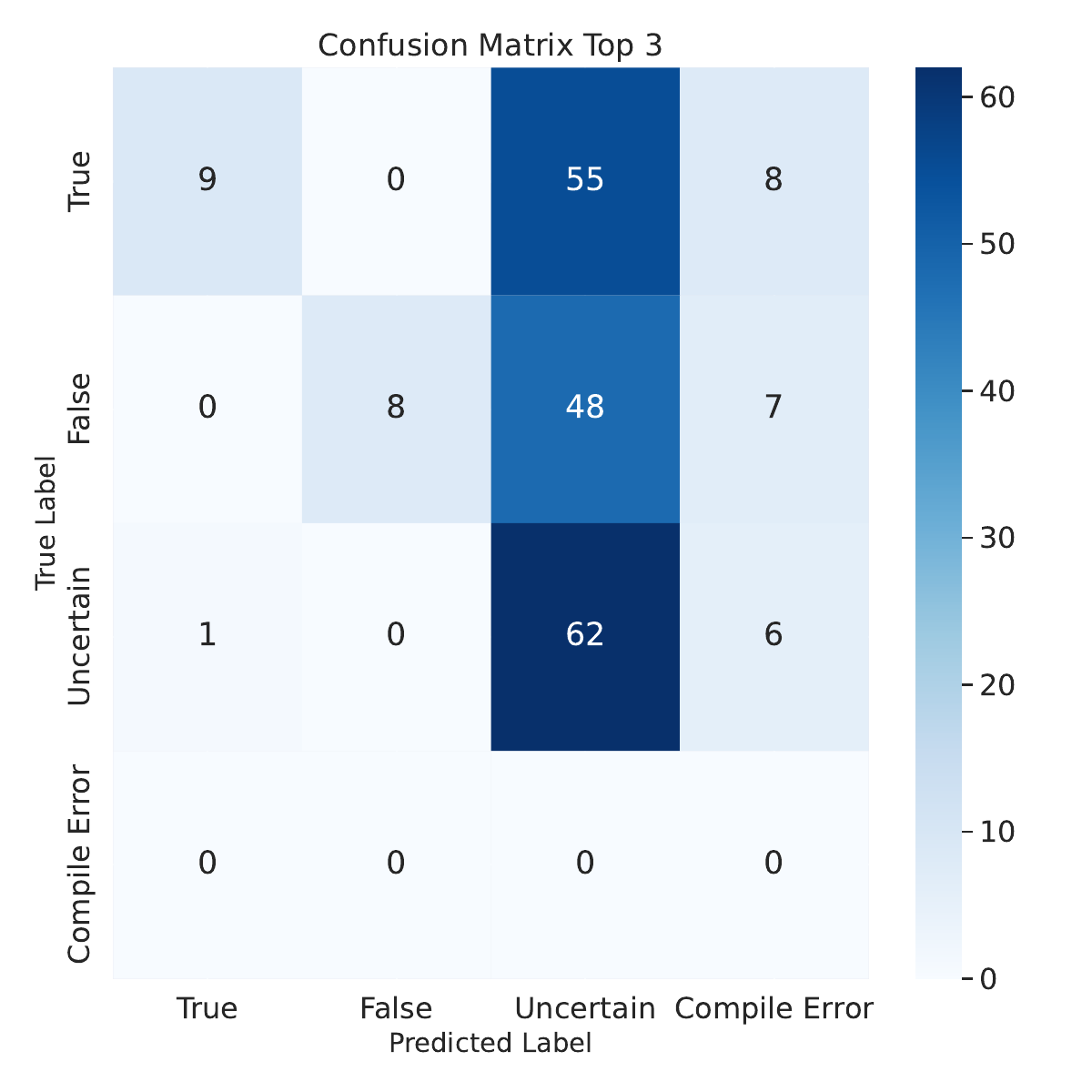}
                \caption{10-Shot \& Top 3}
            \end{subfigure}
            
        \end{minipage}
    }
    \caption{Confusion matrices of our approach (Text-JEPA \& Z3 Solver) on the FOLIO dataset across different shot/top-$k$ settings.}
    \label{fig:confusion_matrix_TextJePA}
\end{figure}

To further investigate the impact of NL2FOL conversion quality on downstream reasoning, we compare confusion matrices of our approach (Text-JEPA + Z3) under different settings: zero-shot, 10-shot (Rank 1), and top-k selections (k = 2, 3). While the zero-shot and Rank 1 settings exhibit similar distributions of uncertain and incorrect reasoning outcomes, the Top-2 and Top-3 settings show slight improvements in the number of correctly reasoned ``True'' labels and reduced uncertainty. However, despite the increase in accuracy, the rise in the ``uncertain'' label poses a challenge. It is difficult to determine whether this uncertainty arises from a lack of strong connections between the premises leading to uncertainty, or if the ``uncertain'' label genuinely corresponds to the correct label. This suggests that in some cases, the top-ranked FOL output may not be the most semantically aligned with the paragraph-level entailment, and leveraging alternative candidates (e.g., via re-ranking or ensembling) could enhance reasoning performance. However, improvements are modest, highlighting the need for better scoring or selection mechanisms beyond confidence ranking alone. For a detailed comparison of the confusion matrices across these settings, please refer to Figure~\ref{fig:confusion_matrix_TextJePA}.

\subsection{Discussion}

Our experiments show that Text-JEPA, despite being based on a compact T5-base model, outperforms larger language models in NL2FOL conversion, while maintaining high stability across prompt variations and domains. Combined with symbolic reasoning, it delivers competitive performance with greater transparency.

However, we find that strong sentence-level conversion performance does not always lead to effective reasoning at the paragraph level. This is because our model converts NL to FOL at the sentence level, while reasoning requires coherent logic across multiple sentences. Moreover, current evaluation compares predicted FOL against gold FOL\_label, which is not available in real-world scenarios—only the original NL input is. This highlights the need for future work to assess conversion quality at the paragraph level, and to develop evaluation protocols that better reflect practical deployment settings, where semantic coherence and structural consistency are crucial. Additionally, our results highlight the implicit nature of LLM-based reasoning. In two-stage settings, SRho-Scores remain low despite high accuracy, suggesting that LLMs may rely on shallow heuristics rather than explicit logic. Symbolic solvers like Z3, while precise on structured inputs, struggle with unstated or implicit knowledge common in natural language.

These findings emphasize the need to improve semantic alignment, discourse-level consistency, and integration of latent knowledge in neuro-symbolic reasoning systems.

\section{Conclusion}

In this work, we introduced Text-JEPA, a lightweight yet effective architecture for converting NL2FOL. Despite its compact T5-base backbone, Text-JEPA consistently outperforms larger LLM-based baselines in conversion accuracy and generalization, while offering more stable and interpretable outputs. When paired with symbolic reasoning using Z3, the model achieves competitive entailment performance, validating the benefits of explicit intermediate representations.

To evaluate the full NL2FOL-to-reasoning pipeline, we proposed a comprehensive framework with three key metrics: Conv-Score to assess symbolic conversion quality, Reasoning-Score for logical inference correctness, and SRho-Score to capture consistency between predicted logic and reasoning outcomes. However, our findings also reveal limitations in current approaches—particularly the gap between sentence-level conversion and paragraph-level reasoning, as well as the reliance on gold FOL labels for evaluation, which are not available in real-world use. These insights point to future directions in improving discourse-level coherence and developing more realistic, NL-grounded evaluation settings for robust neuro-symbolic reasoning.

\bibliographystyle{IEEEtran}
%\bibliography{references.bib}
\bibliography{references}

\end{document}